\documentclass{article}

\newif\ifarxiv
\arxivtrue

\usepackage[preprint,nonatbib]{neurips_2026}
\usepackage[numbers]{natbib}

\usepackage[utf8]{inputenc}
\usepackage[T1]{fontenc}
\usepackage{hyperref}
\usepackage{url}
\usepackage{booktabs}
\usepackage{amsfonts}
\usepackage{amssymb}
\usepackage{amsmath}
\usepackage{nicefrac}
\usepackage{microtype}
\usepackage{xcolor}
\usepackage{colortbl}
\usepackage{enumitem}
\usepackage{graphicx}
\usepackage{subcaption}
\usepackage{wrapfig}
\usepackage{multirow}
\usepackage[disable]{todonotes}
\usepackage{algorithm}
\usepackage{algorithmic}
\usepackage[most]{tcolorbox}

\definecolor{col_evo}{RGB}{52,101,164}
\definecolor{eatink}{HTML}{13161C}
\definecolor{eatnavy}{HTML}{0E1A2B}
\definecolor{eatamber}{HTML}{E8A12E}
\definecolor{eatpaper}{HTML}{F5F3EE}
\definecolor{eatstone}{HTML}{6F737A}
\definecolor{eatline}{HTML}{DEDBD3}
\definecolor{eatsteel}{HTML}{5C7FA3}
\definecolor{eatsand}{HTML}{C9B48A}

\newtcolorbox{supervisorlog}[1][]{
  colback=gray!5, colframe=gray!60, fonttitle=\small\bfseries,
  title=Supervisor Log, boxrule=0.5pt, arc=2pt, left=4pt, right=4pt, top=2pt, bottom=2pt,
  #1
}

\title{Evolving Agent Teams}

\author{%
  Shiyi Cao \quad Ziming Mao \quad Dacheng Li \quad Joseph E. Gonzalez \quad Ion Stoica \\
  University of California, Berkeley \\
  \texttt{\{shicao, ziming.mao, dacheng177, jegonzal, istoica\}@berkeley.edu}
}

\begin{document}

\maketitle

\begin{abstract}
The rapid progress of software engineering agents drives interest in applying them to open-ended problems such as GPU kernel optimization. However, we find that existing coding agents converge to local optima on real-world, expert-written kernels (e.g., \textsc{FlashInfer}).
We introduce \textsc{Evolving Agent Teams}, a multi-agent framework that
escapes such local optima. It features (1) a central supervisor that diagnoses
each local optimum and restructures a team of agents; and (2) a persistent file system that comprises a learning document, an evolution history, and per-team artifacts, grounding restructuring with
accumulated experience.
Under a matched token budget, \textsc{Evolving Agent Teams} consistently
outperforms strong coding-agent baselines. With an extended token budget, \textsc{Evolving Agent Teams} generates a CUDA kernel that reaches $1.2\times$ mean speedup over the production
\textsc{FlashInfer} MLA paged decode kernel from scratch and wins on 36 of 47 production
workloads from \textsc{FlashInfer-Bench} ($0.86$--$0.99\times$ on the remaining 11).
It also demonstrates superior performance on tasks such as KDA prefill on AMD MI350X ($3.01\times$ over vLLM) and NVIDIA B200 ($3.44\times$ over FlashKDA), MoE expert-offload serving on Apple silicon ($30.3$ tokens/s decode for Qwen3.8-Flash-Next in MLX 4-bit on a 48\,GB M4 Pro), and RoboDojo robot manipulation ($62.1$ on held-out seeds vs.\ $55.3$ for the best leaderboard entry).
Code is available at \url{https://github.com/caoshiyi/EAT}.
\end{abstract}

\vspace{-4pt}
\begin{figure}[h!]
\centering
\vspace{-8pt}
\includegraphics[width=0.95\textwidth]{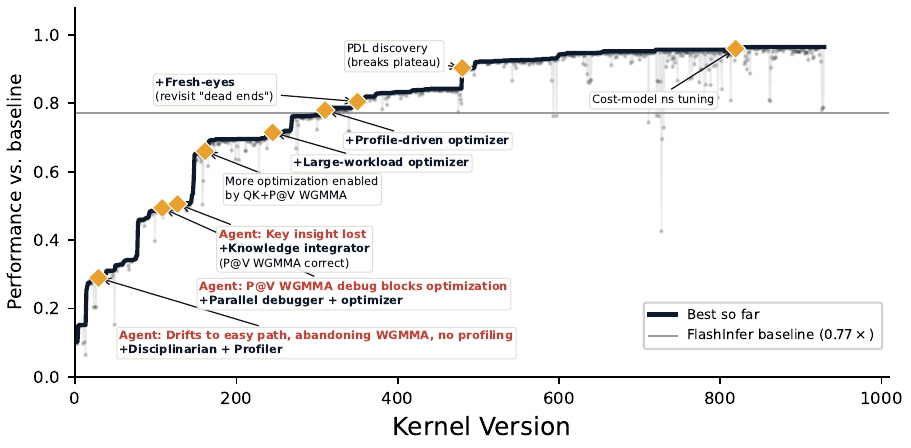}
\vspace{-5pt}
\caption{MLA decode kernel long-running optimization trajectory with Claude Opus~4.6. Diamond markers indicate supervisor interventions that restructure the team. Scores are measured against \textsc{FlashInfer}'s \texttt{run()} stage directly, excluding \texttt{plan()} overhead ($1.0\times$); production \textsc{FlashInfer} solutions cache \texttt{plan()} and only re-execute it when workload metadata changes~\citep{flashinferbench2026} (solid baseline, $0.77\times$). The kernel ultimately reaches $0.96\times$ the plan-excluding baseline, corresponding to $1.2\times$ the production baseline.}
\label{fig:teaser}
\vspace{-10pt}
\end{figure}

\section{Introduction}

The rapid progress of coding agents on software engineering tasks~\citep{jimenez2024swebench, yang2024swe, wang2024openhands} has driven interest in applying them to open-ended discovery tasks such as GPU kernel optimization. Here, the goal is not just to produce a correct
kernel but to search a large, hardware-dependent design space for high performance. Expert-written kernels such as \textsc{FlashInfer}~\citep{ye2025flashinfer} and \textsc{FlashAttention}~\citep{shah2024flashattention3} navigate this space across two levels: low-level primitive use (e.g., WGMMA tensor cores, swizzled shared-memory layouts) and high-level architectural design (e.g., dispatch strategies, warp specialization, dataflow). While low-level primitive use can substantially improve performance within a chosen design, high-level architectural choices often determine which hardware bottlenecks remain and thus largely shape the attainable performance ceiling.

Unfortunately, even strong coding agents struggle to write challenging GPU kernels. They often make rapid initial progress, but then converge to local optima far below expert-written baselines. A common failure mode is that agents select design directions that are initially easier to implement but architecturally limited, spend most of their budget tuning within those directions, and fail to revisit harder alternatives that require substantial debugging or architectural rethinking.

For example, when writing the MLA decode kernel used in DeepSeek-V3 inference~\citep{deepseekv3}, single-agent runs often abandoned difficult WGMMA-based implementations after repeated correctness failures, and instead optimized simpler but less efficient WMMA-based designs. These designs plateaued far below expert-written baselines: Codex 5.5 xHigh reaches only 28\% of expert performance despite a \$1K budget and multiple days of effort. Static agent teams show the same limitation (Section~\ref{sec:results}): a fixed 4-role team built with Claude Code~\citep{anthropic2025claudecode} Opus 4.7 1M reaches only 55\%. These results suggest that neither stronger single agents nor fixed teams are sufficient for challenging GPU kernel optimization. The process moves through distinct phases, from establishing correctness to debugging hardware primitives, profiling bottlenecks, and redesigning the kernel architecture. Each phase requires different expertise, tools, and coordination patterns, causing fixed agent organizations to become mismatched to the current stage.

We introduce \textsc{Evolving Agent Teams}, a multi-agent framework that escapes such local optima by evolving the agent team's organization alongside the optimization process. This evolution is driven by a central supervisor agent, grounded in a persistent file system with accumulated experience. Concretely, the supervisor monitors team progress, reflects on the agent behavior, diagnoses the cause of plateau (e.g., a knowledge gap or an exhausted search), and adapts the team accordingly: spawning specialists for %
high-effort tasks, %
introducing profiling roles to get more insights, or running parallel roles for different directions (Section~\ref{sec:eat}). The file system contains a learning document of empirical observations, per-team artifacts, and an evolution history. %

On the MLA decode kernel, \textsc{Evolving Agent Teams} explores over 100 team configurations. The supervisor repeatedly reconfigures the team around the current failure mode, which improves token efficiency and enables successive architectural advances, including performant WGMMA-based matrix multiplication, a PV accumulation strategy, and L2 prefetching. Together, these improvements allow \textsc{Evolving Agent Teams} to substantially outperform prior single-agent and static-team approaches (Figure~\ref{fig:teaser}).

We evaluate on four challenging CUDA kernel optimization tasks spanning two GPU architectures (Section\ref{sec:results}): Flash Attention~4~\citep{zadouri2026flashattention} and Gated Delta Net~\citep{yang2025gateddeltanet} prefill on Blackwell B200, and MLA paged decode~\citep{deepseekv3, ye2025flashinfer} and fused GEMM+AllReduce (G+AR)~\citep{spector2024thunderkittens} on Hopper H100.
\textsc{Evolving Agent Teams} outperforms strong coding agent baselines on every task. Notably, on the MLA paged decode task, evaluated on 47 production DeepSeek-V3 workloads from \textsc{FlashInfer-Bench}~\citep{flashinferbench2026}, \textsc{Claude Code} agent plateaus at $0.51\times$ the expert-written \textsc{FlashInfer} kernel, while our system reaches $0.74\times$ under a matched token budget.
With additional budget, the system surpasses the production \textsc{FlashInfer} baseline, achieving $1.21\times$ mean speedup
, faster on 36 of 47 workloads, while the previous best result is $0.61\times$ from \textsc{K-Search}~\citep{cao2026k}.

\section{Related Work}

\textbf{Multi-agent LLM systems.}
\textsc{ChatDev}~\citep{qian2024chatdev}, \textsc{MetaGPT}~\citep{hong2024metagpt}, and \textsc{MAGIS}~\citep{tao2024magis} assign fixed roles (programmer, reviewer, tester) to distinct agents for software engineering.
\textsc{AutoGen}~\citep{wu2024autogen} provides a general multi-agent conversation framework; \textsc{DyLAN}~\citep{liu2023dynamic} and \textsc{AgentVerse}~\citep{chen2023agentverse} dynamically adapt the team's role composition during execution, but either select from a hand-crafted pool of role prompts (\textsc{DyLAN}) or regenerate the group from scratch each round without preserving global context across re-compositions (\textsc{AgentVerse}).
Our framework differs in two respects: (1)~it operates over multi-day horizons where the \emph{nature} of productive work changes qualitatively (from knowledge acquisition to search to verification), requiring changes to agent \emph{roles and mandates}, not merely routing; and (2)~the supervisor maintains a persistent global file system across team restructurings.

\textbf{Self-improving agents and evolutionary search.}
\textsc{Reflexion}~\citep{shinn2023reflexion} accumulates verbal feedback across trials;
\textsc{Voyager}~\citep{wang2023voyager} builds a growing skill library.
These approaches improve a single agent's behavior within a fixed role, whereas our framework adapts the broader optimization process around a single agent.
\textsc{FunSearch}~\citep{romera2024funsearch}, \textsc{OpenEvolve}\citep{sharma2025openevolve} and  \textsc{AlphaEvolve}~\citep{novikov2025alphaevolve} use population-based evolutionary search with LLM-driven mutation over programs. Our approach differs in that it employs no population and no crossover; instead, a supervisor performs sequential, diagnosis-driven adaptation of the agent team architecture. \textsc{ADAS}\citep{hu2024adas} uses an LLM with an archive to iteratively propose new agent architectures. Our work also uses an LLM to iteratively adapt the team architecture. However, the purpose of this adaptation is different: rather than searching for a standalone best agent architecture, we evolve the team to better facilitate the optimization process at different stages.

\textbf{Meta-optimization and text optimization.}
A growing body of work uses LLMs to optimize text and code artifacts iteratively.
\textsc{OPRO}~\citep{yang2023large} optimizes prompts conditioned on past (solution, score) pairs;
\textsc{TextGrad}~\citep{yuksekgonul2024textgrad} provides textual feedback on current artifacts;
\textsc{GEPA}~\citep{agrawal2025gepa} uses reflective feedback from rollout traces. \textsc{Meta-Harness}~\citep{lee2026meta} optimizes the code surrounding a fixed LLM by giving a coding-agent proposer full filesystem access to prior candidates' source code, scores, and execution traces. Our work shares the insight that richer access to prior experience enables better meta-optimization. However, the object of optimization is different: Meta-Harness optimizes the surrounding harness, whereas our method adapts the search process itself to better discover high-performance kernel implementations.

\textbf{Code generation and optimization, and inference-time scaling.}
\textsc{SWE-agent}~\citep{yang2024swe}, \textsc{SkyRL-Agent}~\citep{cao2025skyrl} and \textsc{SWE-Gym}~\citep{pan2025swegym} advance agents in software engineering tasks. In GPU kernel optimization, most of the existing LLM-based solutions~\citep{wei2025astra,lange2025towards,liao2025kernelevolve,zhang2025cudaforge} adopt a fixed human-designed system, where LLMs refine kernels based on compilation or profiling feedback. Some recent works~\citep{guo2025evoengineer} add evolutionary strategies into the system for maintaining and selecting candidates.
\textsc{K-Search}~\citep{cao2026k} decouples planning from implementation via a co-evolving world model but employs a fixed two-role configuration.
Inference-time compute scaling~\citep{snell2024scaling,li2025s} improves reasoning within a fixed inference-time search procedure; our framework addresses the complementary problem of adapting the search process itself.

\section{Method}
\label{sec:method}
\begin{figure}[ht!]
\centering
\includegraphics[width=0.95\textwidth]{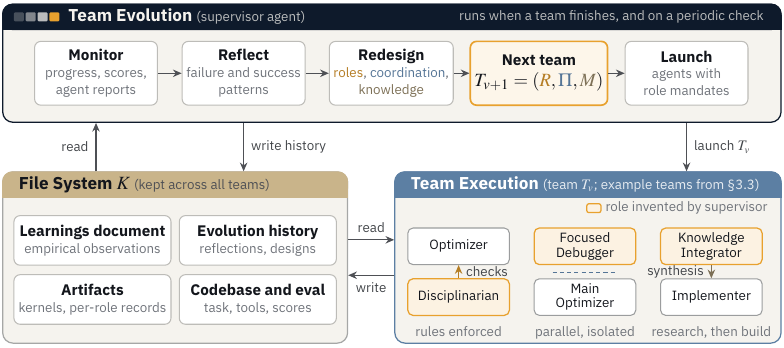}
\caption{The \textsc{Evolving Agent Teams} framework. Team execution runs coding agents on the task through tool use and execution feedback. Team evolution acts as a meta-optimizer: it analyzes the past success and failure patterns, and adapts the team composition, role specialization, coordination protocol, and knowledge management for the current stage of discovery.}
\label{fig:framework}
\end{figure}

\subsection{Problem Setup}
\label{sec:problem}

We consider open-ended program optimization tasks: given a goal description, a codebase, and correctness constraints $\mathcal{C}$, find a solution $s \in \mathcal{P}$ (an executable program) that maximizes a scoring function $\mathcal{F}: \mathcal{P} \to \mathbb{R}$ while satisfying $\mathcal{C}$.
Performance is measured against a finite token budget.

Existing LLM-based systems tackling this problem~\citep{yang2024swe,bui2026building,qian2024chatdev,hong2024metagpt,romera2024funsearch,novikov2025alphaevolve,cemri2026adaevolve} can be characterized by an \emph{agent team configuration} $T = (R, \Pi, M)$ governing three design choices:
\begin{itemize}[leftmargin=*]
  \item \emph{Roles} $R$\,---\,what each agent is responsible for, e.g., a single generalist~\citep{yang2024swe,openai2025codex,anthropic2025claudecode}, fixed programmer/reviewer/tester~\citep{qian2024chatdev,hong2024metagpt}, or an LLM mutation operator~\citep{romera2024funsearch,novikov2025alphaevolve,cemri2026adaevolve}.
  \item \emph{Coordination} $\Pi$\,---\,how agents interact, e.g., no coordination (single agent), a fixed role-to-role pipeline~\citep{qian2024chatdev,hong2024metagpt}, or a select-mutate-evaluate cycle~\citep{romera2024funsearch,novikov2025alphaevolve,cemri2026adaevolve}.
  \item \emph{Knowledge management} $M$\,---\,how experience is accumulated and shared, e.g., the filesystem with no explicit management~\citep{yang2024swe,anthropic2025claudecode}, structured message passing enforced by fixed protocol~\citep{qian2024chatdev,hong2024metagpt}, or a candidate archive~\citep{romera2024funsearch,novikov2025alphaevolve,sharma2025openevolve,cemri2026adaevolve}.
\end{itemize}
In most existing frameworks, all three are human-designed and fixed before the run begins.\footnote{Some evolutionary systems adapt search \emph{hyperparameters} during a run (e.g., AdaEvolve~\citep{cemri2026adaevolve} modulates exploration intensity and budget allocation); here ``fixed'' refers specifically to the team configuration $T = (R, \Pi, M)$, the roles, coordination protocol, and knowledge management strategy, which remain static in all listed systems.}
These systems optimize the solution $s$ but do not adapt $T$ itself, making them prone to stagnation at local optima during long-running open-ended optimization tasks.

\subsection{Evolving Agent Teams}
\label{sec:eat}

\textsc{Evolving Agent Teams} co-evolves $T$ alongside $s$, adapting the optimization system to better facilitate the discovery.
It operates at two levels, following the loop shown in Figure~\ref{fig:framework}: at the \emph{team evolution} level, a supervisor agent monitors each team's execution, analyzes success and failure patterns, and designs the next team $T_{v+1}$.
At the \emph{team execution} level, a team of LLM-based coding agents works on the task under the current configuration $T_v$.
A persistent file system $K$ connects the two levels, accumulating all records across team executions and restructurings: a \emph{learnings document} of accumulated empirical observations, per-team \emph{artifacts} (kernels and per-role records), and the \emph{evolution history} (the supervisor's reflections and design decisions from each version), so that each team can build on and improve upon prior work.

\paragraph{Team Evolution.}
\label{sec:outer-loop}
The supervisor is triggered by two signals: the current team finishing its run, or a fixed-interval cron job. Upon either, it checks the team's progress so far, reflects on any observed failure or success patterns (e.g., drifting to easy paths, key insights from previous teams lost, stuck in debugging, and heavily micro-optimizing), and designs the next team along the three axes of $T$:

\begin{itemize}[leftmargin=*]
  \item \emph{Roles.}
  The supervisor invents roles as needed, rather than selecting from a predefined set.
  A single agent handling implementation, profiling, and research simultaneously divides %
  its limited context across all three, sacrificing depth in each. The supervisor addresses this by creating focused roles such as \textsc{Profiling Scientist} (designs targeted microbenchmarks to isolate bottlenecks), \textsc{Challenger} (stress-tests claimed performance ceilings), \textsc{Primitive Study Agent} (studies potentially useful primitives and their usage, running small tests), and \textsc{Knowledge Integrator} (synthesizes findings scattered across prior teams into actionable plans). Table~\ref{tab:invented-roles} in the appendix lists other observed roles.

  \item \emph{Coordination.}
  The supervisor decides how agents interact: it can pipeline them (research $\to$ implement $\to$ profile), run them in parallel on separate branches, or combine both.
  For example, the supervisor may spawn one agent solely to debug a blocking issue (forbidden from touching the main codebase) while a second agent continues optimizations on the main branch, so that one agent's failure does not block the other's progress.

  \item \emph{Knowledge management.}
  The supervisor evolves how teams manage their experience and learning within the file system $K$, specifying where each agent should output: e.g., write findings into a dedicated document, append learnings during execution to a team-shared log, or commit code to a specific branch.
  For example, when early insights were getting lost across teams, the supervisor introduced a cross-agent synthesis document that connected all prior findings into an actionable plan, and mandated future teams to read it.
\end{itemize}

\paragraph{Team Execution.}
\label{sec:inner}

Each team execution deploys the agents specified by $T_v$ on the task.
Each agent is an LLM-based coding agent~\citep{yang2024swe,bui2026building} with shell access, able to read and write files, execute commands, run compilers and profilers, and iterate autonomously on the codebase.
Agents receive their role mandates and access to the codebase, evaluation tools, and the file system $K$.
One agent's outputs (committed code, profiling reports, written analyses) become inputs for others according to the coordination plan.
\paragraph{File System.}
\label{sec:artifacts}

The file system $K$ is a shared filesystem that persists and grows across all team versions.
$K$ contains the codebase, evaluation tools, and all records produced by prior teams: the \emph{learnings document}, per-team \emph{artifacts} (kernels and per-role records), and the \emph{evolution history}.
This continuity is critical for long-running optimization: each team inherits all prior learnings, failed attempts, and successful strategies, enabling continuous experience accumulation that may open up new optimization directions.

\subsection{MLA Decode: A Case Study in Team Evolution}
\label{sec:mla-case-study}

In this section, we present a detailed case study of the MLA decode kernel optimization with \textsc{Evolving Agent Teams}.

\textbf{Evolution trajectory.}
Figure~\ref{fig:teaser} shows the performance trajectory of the evolving team.
The boxes below show representative evolution steps, each summarizing the observed agent behavior, the supervisor's reflection, the team design change, and the outcome.

\colorlet{eatroles}{eatamber!78!black}
\colorlet{eatcoord}{eatsteel!92!black}
\colorlet{eatknow}{eatsand!62!black}
\newtcbox{\evtag}[1]{on line, colback=#1!13, colframe=#1!13, coltext=#1, arc=1.5pt, boxrule=0pt,
  boxsep=0pt, left=2pt, right=2pt, top=0.2pt, bottom=0pt, fontupper=\sffamily\bfseries\scriptsize}
\newcommand{\evrow}[2]{\par\noindent{\sffamily\bfseries\color{eatnavy}#1}\enspace#2\par}
\newcommand{\evR}{\evtag{eatroles}{Roles}}
\newcommand{\evC}{\evtag{eatcoord}{Coordination}}
\newcommand{\evK}{\evtag{eatknow}{Knowledge}}
\newcommand{\evrule}{\begingroup\setlength{\fboxsep}{0pt}%
  \textcolor{eatpaper!25!eatnavy}{\rule{0.5em}{0.45em}}\,\textcolor{eatpaper!50!eatnavy}{\rule{0.5em}{0.45em}}\,%
  \textcolor{eatpaper!75!eatnavy}{\rule{0.5em}{0.45em}}\,\textcolor{eatamber}{\rule{0.5em}{0.45em}}\endgroup}
\newtcolorbox{evolbox}[3]{
  enhanced, colback=white, colframe=eatline!80!eatstone, boxrule=0.45pt, arc=2.5pt,
  colbacktitle=eatnavy, coltitle=eatpaper, fonttitle=\sffamily\footnotesize,
  titlerule=0pt, title={\evrule\quad\textbf{#1}\quad{\color{eatpaper!78!eatnavy}#2}\hfill\textbf{#3}},
  left=5pt, right=4pt, top=1.5pt, bottom=0.5pt, toptitle=0.8pt, bottomtitle=0.6pt, lefttitle=5pt, righttitle=5pt,
  fontupper=\footnotesize, before upper={\parskip=0pt}, before skip=5pt plus 3pt, after skip=5pt plus 3pt,
}

\begin{evolbox}{V1 $\to$ V2}{Enforcing discipline}{0.15 $\to$ 0.065 \textmd{\color{eatpaper!78!eatnavy}(first correct WGMMA kernel)}}
\evrow{Behavior}{Solo agent drifted from WGMMA to scalar code under pressure; no profiling was done; multi-variable changes made attribution impossible.}
\evrow{Reflection}{The agent needs external discipline. Without enforcement, it abandons hard constraints when they become inconvenient.}
\evrow{Design}{\evR{} +Disciplinarian (blocks non-WGMMA commits, enforces one-change-per-commit). \evC{} Disciplinarian checks after every commit. \evK{} Introduced \texttt{LEARNINGS.md} for persistent cross-run discoveries.}
\evrow{Outcome}{First validated correct WGMMA kernel. Discipline held (one rule-gaming attempt blocked by supervisor).}
\end{evolbox}

\begin{evolbox}{V5 $\to$ V6}{Supervisor-orchestrated parallelism}{0.34 $\to$ 0.49 \textmd{\color{eatpaper!78!eatnavy}(+44\%)}}
\evrow{Behavior}{V3--V5 each independently attempted P@V WGMMA integration and failed on the same root cause. Structural changes require isolation debugging then integration --- one agent session is too short for both.}
\evrow{Reflection}{Offering pair-spawning to the agent doesn't work (V5 never used it). The supervisor should orchestrate parallelism directly.}
\evrow{Design}{\evR{} Focused Debugger (P@V WGMMA only, forbidden from main kernel) + Main Optimizer (micro-ops only, forbidden from P@V). \evC{} Fully parallel, isolated branches.}
\evrow{Outcome}{Debugger failed P@V conclusively (dead end confirmed). Optimizer made 12 commits for +44\%, unblocked by the parallel pattern.}
\end{evolbox}

\begin{evolbox}{V7 $\to$ V8}{Cross-team knowledge synthesis}{0.50 $\to$ 0.51 \textmd{\color{eatpaper!78!eatnavy}(knowledge breakthrough)}}
\evrow{Behavior}{V4 had recommended ``use WGMMA RS with P in registers'' but V5--V7 all ignored this because the recommendation was in V4's report, not in the shared learnings.}
\evrow{Reflection}{Lost knowledge between teams is the biggest structural weakness. Per-experiment facts are captured but cross-experiment strategic recommendations are not.}
\evrow{Design}{\evR{} +Knowledge Integrator. \evK{} Introduced \texttt{SYNTHESIS.md} connecting all agents' findings into an actionable plan. Mandated all future teams to read it.}
\evrow{Outcome}{RS standalone test passed (first P@V WGMMA success ever). Directly enabled V9's integration milestone.}
\end{evolbox}

\begin{evolbox}{V9 $\to$ V10}{Re-profiling after a structural change}{0.50 $\to$ 0.66 \textmd{\color{eatpaper!78!eatnavy}(+32\%)}}
\evrow{Behavior}{V9 integrated P@V WGMMA RS, changing the kernel's bottleneck structure completely.}
\evrow{Reflection}{With both GEMMs on WGMMA, the kernel needs fresh profiling --- all prior bottleneck analysis is stale.}
\evrow{Design}{\evR{} Standard Profile-Driven Optimizer on the new architecture. \evC{} Solo, profile-first discipline. \evK{} Inherited full $K$ including V9's integration learnings.}
\evrow{Outcome}{9 commits, +32\%. Split-K retune alone gave +4.1. Structural change unlocked entirely new optimization surface.}
\end{evolbox}

This trajectory illustrates the strength of \textsc{Evolving Agent Teams} in responding to failures that no fixed design can foresee: the supervisor observes a concrete dysfunction, diagnoses its structural cause, and restructures roles, coordination, or knowledge management to address it. 

Section~\ref{sec:results} evaluates how these adaptations translate into quantitative gains across multiple tasks.

\section{Experiments}
\label{sec:results}

\begin{table}[th!]
\centering
\caption{Settings of Main Evaluation Tasks.}
\label{tab:tasks}
\small
\resizebox{0.9\textwidth}{!}{%
\begin{tabular}{lllcll}
\toprule
\textbf{Task} & \textbf{GPU} & \textbf{Language} & \textbf{Workloads} & \textbf{Baseline} & \textbf{Seed} \\
\midrule
FA4 & B200 & CuTe DSL & 24 & FlashAttention-4~\citep{zadouri2026flashattention} & $0.17\times$ CuTe DSL kernel \\
GDN~\citep{yang2025gateddeltanet} & B200 & CUDA & 100 & FlashInfer~\citep{ye2025flashinfer} & PyTorch reference \\
MLA~\citep{deepseekv3} & H100 & CUDA & 47 & FlashInfer~\citep{ye2025flashinfer} & PyTorch reference \\
G+AR & H100 & CUDA & 5 & CUTLASS~\citep{thakkar2023cute}+NCCL & PyTorch reference \\
\bottomrule
\end{tabular}%
}
\end{table}

\subsection{Experimental Setup}

\textbf{Tasks.}
We evaluate on four GPU kernel optimization tasks spanning two architectures (Table~\ref{tab:tasks}).
\emph{FA4} targets the attention forward pass~\citep{zadouri2026flashattention}, written in CuTe DSL~\citep{cecka2026cute}; the challenge is co-designing algorithm and kernel pipelining for Blackwell's asymmetric hardware scaling, where tensor core throughput doubles but shared memory bandwidth and non-matmul units do not.
\emph{GDN} computes a prefill pass for Gated DeltaNet~\citep{yang2025gateddeltanet}, a linear RNN with a matrix-valued hidden state; the challenge is parallelizing the inherently sequential recurrence over long sequences via a chunkwise algorithm.
\emph{MLA} implements paged decode for DeepSeek-V3's Multi-head Latent Attention~\citep{deepseekv3}; the challenge is overlapping CUDA core and tensor core work efficiently under extreme register pressure due to MLA's large head dimension.
\emph{G+AR} fuses GEMM with inter-GPU AllReduce into a single kernel, overlapping computation and communication; efficient fusing requires writing a custom WGMMA GEMM kernel with tile-level scheduling to overlap reduction with compute.

\textbf{Metrics.}
For each task, we report the speedup relative to the production baseline: the geometric mean of $t_{\text{base}} / t_{\text{kernel}}$ across a set of workloads, where $t_{\text{base}}$ and $t_{\text{kernel}}$ are the baseline and optimized kernel latencies respectively.
Values above $1.0\times$ indicate the optimized kernel is faster than the baseline; values below $1.0\times$ indicate it is slower.
All settings are run twice; we report the best result from each pair.
We acknowledge this limits statistical conclusions and discuss implications in Section~\ref{sec:conclusion}.

\textbf{Workloads.}
GDN prefill and MLA workloads are drawn from \textsc{FlashInfer-Bench}~\citep{flashinferbench2026}, which provides representative production serving shapes.
FA4 workloads use the same shapes as used in the performance report in the FlashAttention-4 paper~\citep{zadouri2026flashattention}.
GEMM+AR (G+AR) workloads are taken from the ThunderKittens benchmark suite~\citep{spector2024thunderkittens}.

\textbf{Baselines.}
We compare against three baseline systems.
(1)~\emph{Claude Code}~\citep{anthropic2025claudecode}: \textsc{Claude Code} powered by Claude Opus 4.6 with 1M context window, a state-of-the-art coding agent with file editing, shell access, and compiler invocation that autonomously writes, compiles, profiles, and iterates on kernel code.
It receives the same task description, reference kernel, and test harness as the evolving team.
Evaluated on all four tasks in the budget-controlled experiment (Section~\ref{sec:progress-curves}).
(2)~\emph{Fixed four-role team}: a static team of four common high-value roles for GPU kernel optimization (Profiler, Hardware Researcher, Planner, Implementer), each powered by Claude Code with the same Opus 4.7 1M model, but with no supervisor and no team restructuring.
Evaluated on MLA in the ablation study (Section~\ref{sec:ablation}).
(3)~\emph{Codex}~\citep{openai2025codex}: a strong coding agent powered by GPT-5.5 xhigh with 258K context.
Evaluated on MLA in the ablation study (Section~\ref{sec:ablation}).
(4)~\emph{Kernel-specific systems}: \textsc{K-Search}~\citep{cao2026k} and \textsc{CUDAForge}~\citep{zhang2025cudaforge}, run from their publicly released implementations with Claude Opus 4.6.
Evaluated on MLA (Section~\ref{sec:kernel-baselines}).
For Claude Code, \textsc{K-Search}, \textsc{CUDAForge}, and \textsc{Evolving Agent Teams}, we run each setting twice and report the best result; the fixed-team and Codex ablations are single runs.

\textbf{Implementation.}
\textsc{Evolving Agent Teams} is built on top of \textsc{Claude Code}~\citep{anthropic2025claudecode} agents powered by Claude Opus 4.6 with 1M context.
The supervisor is itself a Claude Code agent that orchestrates team evolution by spawning and monitoring worker agents.
There are no predefined initial roles or team configurations; the supervisor designs all roles, coordination plans, and knowledge management strategies from scratch based on the task description.

\subsection{Budget-Controlled Progress Curves}
\label{sec:progress-curves}

\begin{figure}[t]
\centering
\begin{subfigure}[b]{\textwidth}
\centering
\includegraphics[width=\textwidth]{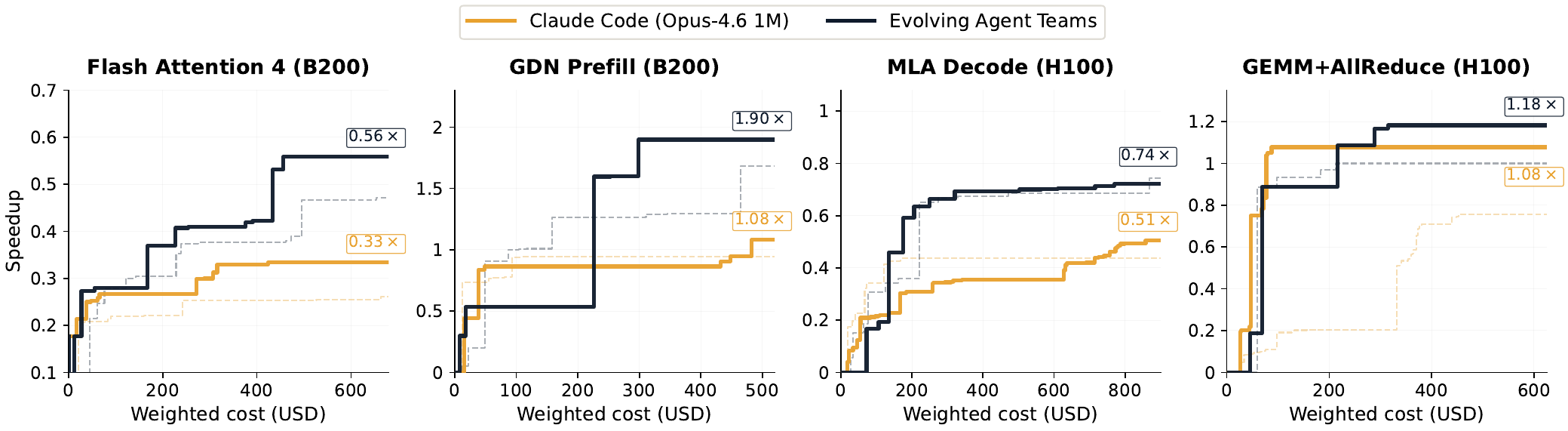}
\caption{Budget-controlled progress curves.}
\label{fig:progress-curves}
\end{subfigure}\\[4pt]
\begin{subfigure}[b]{\textwidth}
\centering
\includegraphics[width=\textwidth]{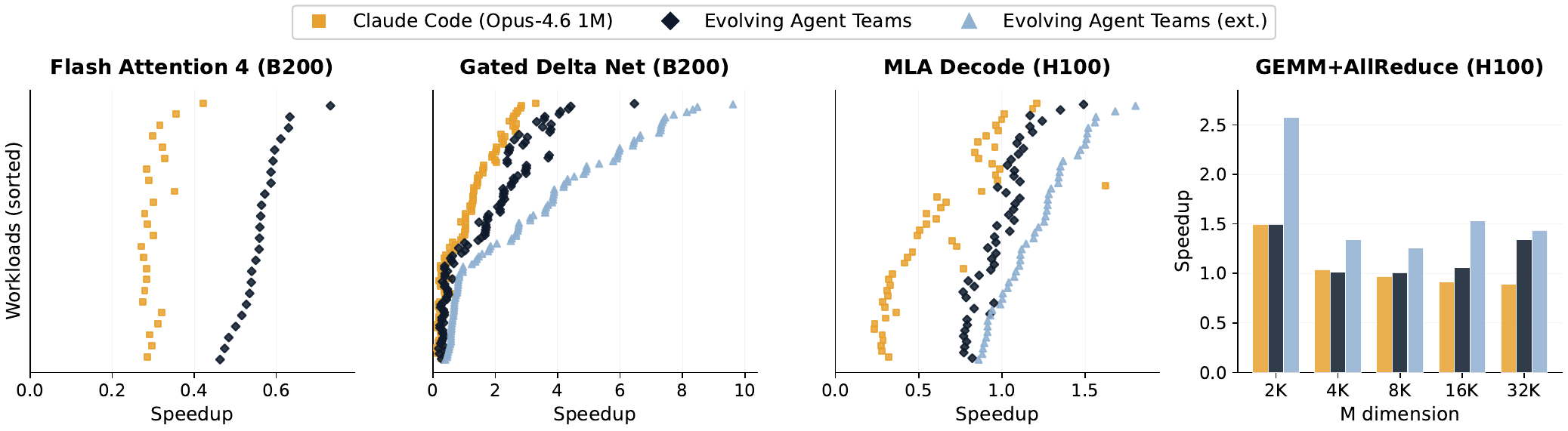}
\caption{Per-workload speedups.}
\label{fig:per-workload}
\end{subfigure}
\caption{Performance comparison across four kernel tasks. (a)~Solid lines show the best of two runs; dashed lines show the second run. Cost computed under published per-token pricing (Claude Opus 4.6 1M). (b)~Each dot is one workload. ``Evolving Teams (ext.)'' denotes longer runs beyond the budget-controlled setting; \textsc{Evolving Agent Teams} shifts the distribution rightward on all tasks.}
\label{fig:progress}
\vspace{-5mm}
\end{figure}
Figure~\ref{fig:progress-curves} plots performance vs.\ cumulative weighted cost, while Figure~\ref{fig:per-workload} shows per-workload speedup distributions.
Cost is computed from per-token API pricing: for Claude Opus 4.6 (1M context), we use \$0.50 / \$6.25 / \$5.00 / \$25.00 per million tokens for cache-read / cache-create / input / output respectively; for GPT-5.5 xhigh (Codex ablation), \$0.50 / \$5.00 / \$30.00 per million tokens for cached-input / input / output.

\textsc{Evolving Agent Teams} outperforms Claude Code: Claude Code curves flatten early on FA4 and GDN Prefill, while \textsc{Evolving Agent Teams} continues improving with more budget spent.
On MLA decode, Claude Code plateaus around $0.44$--$0.51\times$: it abandons the difficult WGMMA tensor core implementation after repeated failures at correctness and instead uses the WMMA API, which limits the tensor core utilization compared to WGMMA.
The evolving team reaches $0.75\times$ under the budget-controlled setting ($1.21\times$ with additional budget, as shown in Figure~\ref{fig:teaser}).

The GDN prefill panel shows particularly high speedups ($>2\times$) because the 100-workload benchmark includes many small input sizes where FlashInfer's planning and preprocessing overhead dominates end-to-end latency, even with \texttt{plan()} cached across calls.
Our generated kernel requires no planning stage, so on these small workloads it achieves disproportionately large speedups. Under the budget-controlled setting, the best \textsc{Evolving Agent Teams} run has a worst workload of $0.19\times$ with 43 of 100 below the baseline, compared to $0.13\times$ and 49 of 100 for the best Claude Code run.
With additional budget (Section~\ref{sec:extended}), the worst workload improves to $0.40\times$ with 37 of 100 below the baseline.

On GEMM+AllReduce, the single-agent run produces a cuBLAS-based kernel, yielding a mean speedup of $1.08\times$ compared to the cuBLAS + NCCL baseline. The budget-controlled evolving team similarly converges on a cuBLAS+NCCL pipeline, achieving $1.18\times$ with stronger gains at larger matrix sizes (32K: $1.34\times$). With additional budget (Section~\ref{sec:extended}), the team discovers a fused WGMMA+NVLS kernel that bypasses cuBLAS entirely and leverages NVLink multicast for the all-reduce, reaching $1.49\times$ overall and $2.57\times$ on the smallest workload where communication latency dominates.

Across all four tasks, the best \textsc{Evolving Agent Teams} kernel matches or exceeds the best Claude Code kernel on nearly every workload (Figure~\ref{fig:per-workload}), with a single exception on MLA decode.
This broad coverage suggests the system does not overfit to a narrow regime; we attribute it to the supervisor's tendency to evolve roles such as \textsc{Workload Analyst} that explicitly diagnose per-workload regressions and steer optimization toward balanced performance (Appendix~\ref{app:invented-roles}).

\subsection{Scaling with Additional Budget}
\label{sec:extended}

\begin{figure}[th!]
\centering
\begin{subfigure}[b]{0.5\textwidth}
\centering
\includegraphics[width=\textwidth]{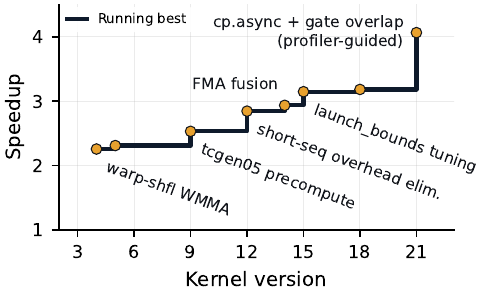}
\caption{GDN Prefill (B200).}
\label{fig:ext-gdn}
\end{subfigure}\hfill
\begin{subfigure}[b]{0.5\textwidth}
\centering
\includegraphics[width=\textwidth]{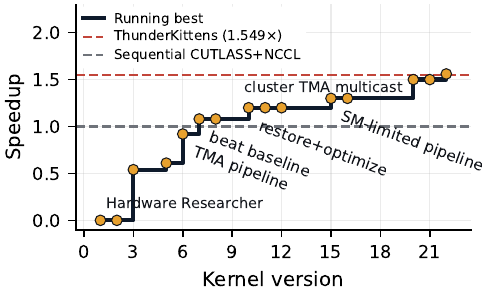}
\caption{GEMM+AllReduce (H100).}
\label{fig:ext-gar}
\end{subfigure}
\caption{Evolution trajectories for GDN and G+AR with additional budget beyond the controlled comparison. Each point is one team version. The step line tracks the running best.}
\label{fig:ext-trajectories}
\end{figure}

Beyond the budget-controlled comparison, we ran \textsc{Evolving Agent Teams} with additional token budget on GDN and G+AR (Figure~\ref{fig:ext-trajectories}).

\textbf{GDN Prefill} (Figure~\ref{fig:ext-gdn}) continues to improve from a $2.2\times$ kernel to $4.06\times$ mean speedup across 21 team versions. With the additional budget, the worst-case workload improves $0.40\times$, and the number of workloads below the baseline drops from 43 to 37 of 100, indicating that extended optimization disproportionately benefits the hardest workloads where the initial kernel architecture was less suited.

\textbf{G+AR} (Figure~\ref{fig:ext-gar}) reaches $1.56\times$ over the cuBLAS+NCCL baseline, comparable to ThunderKittens~\citep{spector2024thunderkittens}, which is at $1.54\times$ over cuBLAS+NCCL.
The trajectory illustrates three distinct phases that require qualitatively different capabilities.
Versions 1--5 focus on correctness: the team debugs WGMMA register layouts and IPC cache coherence for multi-GPU communication, producing a working but slow two-phase kernel ($0.54\times$).
Versions 6--10 introduce TMA asynchronous loads and a warp-specialized pipeline that first exceeds the baseline ($1.08\times$).
Versions 15--22 shift to a fundamentally different architecture: the team replaces cuBLAS with a custom WGMMA kernel fused with NVLink multicast for the all-reduce, culminating in cluster-level TMA multicast ($1.56\times$).

\subsection{Optimizing an Already-Optimized Implementation}
\label{sec:flash-kmeans}

\begin{figure}[t]
\centering
\begin{subfigure}[b]{0.48\textwidth}
\centering
\includegraphics[width=\textwidth]{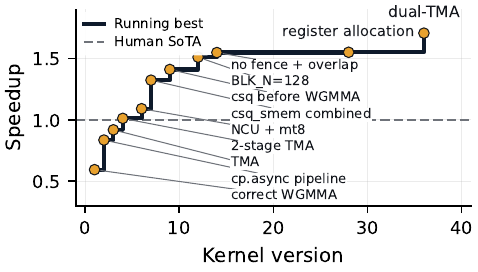}
\caption{Evolution trajectory.}
\label{fig:ext-fkm}
\end{subfigure}\hfill
\begin{subfigure}[b]{0.48\textwidth}
\centering
\includegraphics[width=\textwidth]{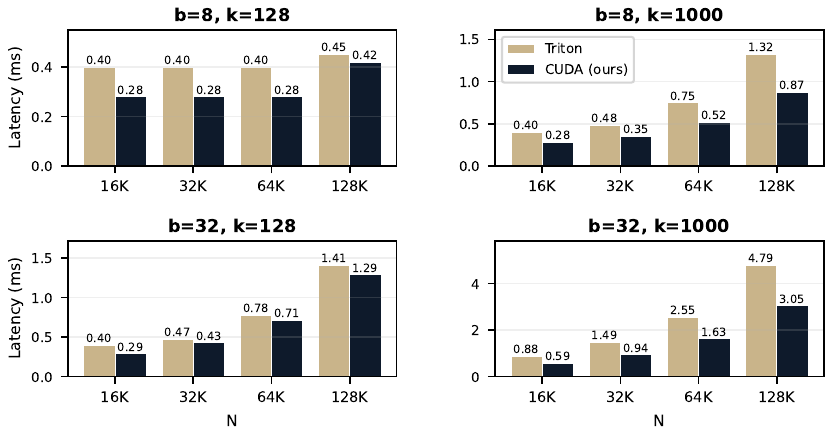}
\caption{Benchmark grid: latency comparison (ours vs.\ Triton).}
\label{fig:flash-kmeans-bench}
\end{subfigure}
\caption{Flash-KMeans (H100). (a)~Evolution trajectory; each point is one kernel version. (b)~Per-iteration latency comparison across the paper's benchmark grid ($D{=}128$, FP16). Our CUDA WGMMA kernel achieves up to $1.5\times$ lower latency on compute-bound configurations (high $K$), while matching Triton on memory-bound regimes.}
\label{fig:flash-kmeans}
\end{figure}

The four tasks above start from na\"ive reference kernels.
To test whether \textsc{Evolving Agent Teams} can also improve code that has already been carefully optimized by domain experts, we apply it to \emph{Flash-KMeans}~\citep{yang2026flash}, a recent GPU-optimized exact K-Means implementation.
Flash-KMeans fuses distance computation with online argmin (FlashAssign) to bypass HBM materialization of the $N{\times}K$ distance matrix and introduces a sort-inverse update that transforms high-contention atomic scatters into segment-level localized reductions, achieving up to $17.9\times$ end-to-end speedup over prior GPU baselines on H200.
The released implementation uses hand-tuned Triton kernels.
We task the \textsc{Evolving Agent Teams} with rewriting the assignment kernel in CUDA, starting from the Triton source as a reference rather than a na\"ive PyTorch baseline.
As shown in Figure~\ref{fig:flash-kmeans}, the system produces a CUDA WGMMA kernel that reaches $1.71\times$ speedup over the Triton kernel on the default workload, and achieves up to $1.5\times$ lower latency on compute-bound configurations (high $K$) across the full benchmark grid (Figure~\ref{fig:flash-kmeans-bench}).

\subsection{Ablation Studies}
\label{sec:ablation}

\begin{wrapfigure}{r}{0.48\textwidth}
\vspace{-12pt}
\centering
\includegraphics[width=0.48\textwidth]{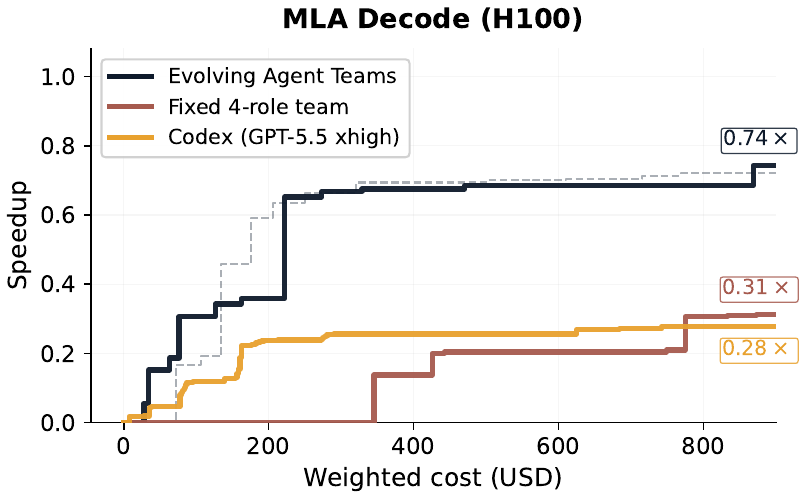}
\caption{Ablation on MLA decode.}
\label{fig:ablation}
\vspace{-10pt}
\end{wrapfigure}

Figure~\ref{fig:ablation} compares \textsc{Evolving Agent Teams} against two ablation baselines on MLA decode.
The \emph{fixed four-role team} (Profiler, Hardware Researcher, Planner, Implementer) reaches $0.31\times$: role specialization helps early on, but the static configuration cannot adapt when the optimization landscape shifts, e.g., it lacks a \textsc{Knowledge Integrator} to prevent cross-agents information asymmetry or a \textsc{Challenger} to break premature ceiling declarations (examples provided in Appendix~\ref{app:fixed-team-case-study}).
\emph{Codex} (GPT-5.5 xhigh, 258K context) reaches $0.28\times$, confirming that the gains of \textsc{Evolving Agent Teams} ($0.74\times$) are not solely due to model choice.
The $0.31\times \to 0.74\times$ gap between the fixed team and the evolving team isolates the value of team evolution: both use the same model and multi-agent structure, but only the evolving system can restructure roles and coordination in response to diagnosed failure modes.

\subsection{Comparison with Kernel-Specific Systems}
\label{sec:kernel-baselines}

\begin{table}[t]
\centering
\caption{Comparison with kernel-specific optimization systems on MLA decode (H100) under a matched \$800 model-inference budget. All kernels are re-evaluated on the same H100 across the 47 production workloads and pass all 47 correctness tests. Speedup is relative to production \textsc{FlashInfer}.}
\label{tab:kernel-baselines}
\small
\begin{tabular}{lccc}
\toprule
\textbf{System} & \textbf{Run 1} & \textbf{Run 2} & \textbf{Best-kernel MMA path} \\
\midrule
\textsc{CUDAForge}~\citep{zhang2025cudaforge} & $0.448\times$ & $0.283\times$ & scalar/SIMT (0/2 WGMMA) \\
\textsc{K-Search}~\citep{cao2026k} & $0.602\times$ & $0.598\times$ & WMMA (0/2 WGMMA) \\
\textsc{Evolving Agent Teams} & $\mathbf{0.721\times}$ & $\mathbf{0.685\times}$ & WGMMA (2/2) \\
\bottomrule
\end{tabular}
\end{table}

We further compare against two systems designed specifically for GPU kernel optimization, \textsc{K-Search}~\citep{cao2026k} and \textsc{CUDAForge}~\citep{zhang2025cudaforge}, on MLA decode.
All systems use Claude Opus 4.6 with medium thinking effort and the same \$800 model-inference budget, and each is run twice.
As shown in Table~\ref{tab:kernel-baselines}, \textsc{Evolving Agent Teams} outperforms both systems in each run.
The gap is explained by the implementation family each system converges to: \textsc{Evolving Agent Teams} produces WGMMA-based kernels in both runs, whereas \textsc{K-Search} converges to WMMA and \textsc{CUDAForge} to scalar/SIMT code in both runs.
As with Claude Code (Section~\ref{sec:progress-curves}), without restructuring the search they settle on an easier-to-get-correct implementation.

\subsection{Surpassing State-of-the-Art Systems}
\label{sec:generalization}

\begin{figure}[t]
\centering
\includegraphics[width=\textwidth]{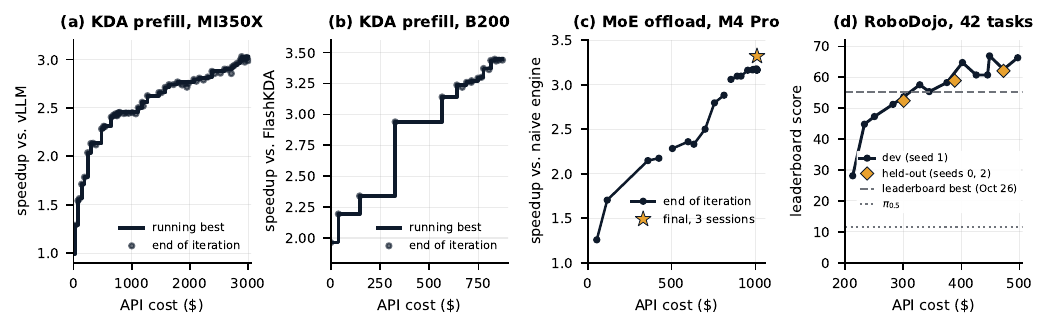}
\vspace{-5mm}
\caption{Full runs on four further settings: harness evaluation at the end of each iteration vs.\ API cost. (a)~MI350X, from a port of vLLM's kernels. (b)~B200, from a port of \textsc{FlashInfer}'s KDA kernels ($1.96\times$). (c)~MoE offload from the naive engine; star: final engine on the recorded sessions. (d)~RoboDojo, 42 tasks: dev (seed~1) and held-out (seeds~0, 2) scores, with leaderboard references as of Oct.~26, 2026.}
\label{fig:new-domains}
\vspace{-3mm}
\end{figure}

Finally, we ask whether full runs can surpass the best existing systems on new hardware and in new domains, with the same supervisor prompt, no predefined roles, and newer models (Opus/Sonnet~5.5; single runs; details in Appendix~\ref{app:generalization}, solo ablations in Appendix~\ref{app:solo-ablations}).
On Kimi Delta Attention (KDA) prefill~\citep{kimi2025linear}, the team's kernel is $3.01\times$ faster than vLLM's hand-written production HIP kernels~\citep{kwon2023vllm} on AMD MI350X, and $3.44\times$ faster than FlashKDA~\citep{flashkda2026}, a fused CUTLASS implementation, on B200 (Figure~\ref{fig:new-domains}a,b). On B300, the final kernel scores $3.54\times$ in the same harness, against $3.02\times$ for the agent-written KDA$^2$ kernels~\citep{kda2_2026} (Appendix~\ref{app:generalization}).
It makes a Metal port of FreeToken~\citep{yang2026freetoken}, which runs Qwen3.8-Flash-Next in MLX 4-bit on a 48\,GB Apple M4 Pro by streaming experts from SSD, serve recorded coding-agent sessions $3.32\times$ faster (Figure~\ref{fig:new-domains}c).
On RoboDojo~\citep{chen2026robodojo} it writes a code policy for 42 simulated dual-arm manipulation tasks that scores $62.1$ on held-out seeds, above the best leaderboard entry ($55.3$ as of Oct.~26, 2026) and the $\pi_{0.5}$ VLA~\citep{black2025pi05} alone ($11.4$) (Figure~\ref{fig:new-domains}d).
The B200 and RoboDojo runs had no web access, and the MoE run had web tools disabled.

\section{Conclusion}
\label{sec:conclusion}

Strong coding agents make rapid initial progress on GPU kernel optimization but converge to local optima well below expert-written baselines.
\textsc{Evolving Agent Teams} escapes these local optima by evolving the team's organization alongside the search: a central supervisor diagnoses each plateau and restructures the team's roles, grounded in a persistent file system. Across four CUDA kernel optimization tasks on Hopper and Blackwell GPUs, it consistently outperforms strong coding-agent baselines under a matched token budget, and with additional budget it surpasses the production \textsc{FlashInfer} MLA decode kernel at $1.21\times$ mean speedup across 47 workloads. It also demonstrates superior performance on tasks such as KDA prefill on AMD and NVIDIA GPUs, MoE expert-offload serving on Apple silicon, and robot manipulation.

\textbf{Limitations.}
We run each main setting twice and each setting in Section~\ref{sec:generalization} once, which limits statistical comparisons; a cost-vs-performance Pareto analysis and a controlled study of noisier domains remain future work.

\newpage
\begin{ack}
This work was done at the Sky Computing Lab. We gratefully acknowledge compute support from SemiAnalysis, Amazon, AMD, and NVIDIA. Shiyi Cao and Dacheng Li are supported by the Amazon Fellowship.
\end{ack}
\bibliographystyle{abbrvnat}
\bibliography{references}

\newpage
\appendix
\section{Observed Roles Invented by the Supervisor}
\label{app:invented-roles}

Table~\ref{tab:invented-roles} lists the roles invented by the supervisor across our experiments.
None of these roles were predefined; the supervisor created each in response to a specific diagnosed need.

\begin{table}[h]
\centering
\caption{Roles invented by the supervisor, grouped by category. All roles emerged from the supervisor's reflection on team performance.}
\label{tab:invented-roles}
\small
\setlength{\extrarowheight}{2pt}
\begin{tabular}{lp{9cm}}
\toprule
\textbf{Role} & \textbf{Description} \\
\midrule
\\[-6pt]
\rowcolor{eatnavy!8} \multicolumn{2}{l}{\textbf{\textsf{Implementation \& Optimization}}} \\[3pt]
Solo Optimizer & Full-stack agent with complete autonomy over design, implementation, and profiling \\
Focused Optimizer & Targets a specific subsystem or bottleneck (e.g., Large-Workload, Memory, L1 Cache, Barrier-Free Reduce, Stage Pipelining) \\
Focused Debugger & Scoped to a single standalone test file; forbidden from touching the main kernel \\
Focused Integrator & Integrates a specific validated component with corrected context \\
Fresh-Eyes Agent & Retries previously dismissed approaches without preconceptions \\[4pt]
\rowcolor{eatsteel!16} \multicolumn{2}{l}{\textbf{\textsf{Analysis \& Profiling}}} \\[3pt]
Profile-Driven Optimizer & Mandated to run \texttt{ncu} before every change and cite profiling data \\
Workload Analyst & Per-workload performance breakdown to identify the biggest gap \\
Compiler Specialist & Investigates compiler behavior (\texttt{ptxas}, SASS, register allocation) \\
Phase Profiler & Instruments kernel with cycle counters to measure per-phase costs \\
Stall Localizer & \texttt{ncu --set source} per-PC profiling to attribute stalls to code lines \\
Roofline Analyst & Computes mathematical upper bounds on achievable performance \\[4pt]
\rowcolor{eatamber!16} \multicolumn{2}{l}{\textbf{\textsf{Knowledge \& Research}}} \\[3pt]
Knowledge Integrator & Synthesizes findings from prior sessions into a single actionable document \\
Dead-End Revisitor & Re-evaluates all revisit log entries against current kernel state \\
Research Optimizer & Web search to study external implementations for structural insights \\
Hardware Cataloguer & Builds and maintains a catalog of available hardware primitives \\
Primitive Study Agent & Studies potentially useful primitives and runs small validation tests \\[4pt]
\rowcolor{eatsand!28} \multicolumn{2}{l}{\textbf{\textsf{Quality Assurance}}} \\[3pt]
Disciplinarian & Enforces constraints (WGMMA, one-change-per-commit, profiling-first) \\
Contrarian / Challenger & Questions assumptions and stress-tests claimed performance ceilings \\
Auditor & Cross-checks proposed designs against known dead ends and hardware constraints \\[4pt]
\rowcolor{eatline!70} \multicolumn{2}{l}{\textbf{\textsf{Architecture \& Design}}} \\[3pt]
Architect & Proposes new kernel designs from a blank slate based on hardware constraints \\
Schedule Explorer & Cost model to systematically enumerate and rank parallelization strategies \\
DAG Topology Explorer & Analyzes the kernel's dataflow graph to identify overlapping opportunities \\
\bottomrule
\end{tabular}
\setlength{\extrarowheight}{0pt}
\end{table}

\section{Case Study: Why the Fixed Four-Role Baseline Starts Slowly}
\label{app:fixed-team-case-study}

Figure~\ref{fig:ablation} shows the fixed four-role baseline (Planner, Hardware Researcher, Implementer, and Profiler; all Claude Opus 4.7, coordinating through a shared filesystem and a SendMessage primitive) reaching its first correctness-passing commit substantially later than the single agent on MLA decode.
Most of this delay is traceable to a single class of bug: a low-level SMEM-layout mismatch in which WGMMA on SM90 requires $8\times64$ atom-tiled shared memory, but the Implementer wrote row-major tiles, silently corrupting the QK output.

The information needed to identify the bug is present in the team but partitioned across agents: the Hardware Researcher holds the layout requirement, the Implementer holds the actual SMEM declaration, and both agree on the descriptor numerics, leading them to localize the bug elsewhere.
The team iterates through incorrect hypotheses without any agent reading the Implementer's source to inspect the layout, despite all having Read access.
A single agent holding both specification and code in its context would identify the discrepancy on direct inspection.

The failure is structural rather than capability-driven.
Each role's charter implicitly scopes its outputs (the Hardware Researcher operates as a recipe expert, not an auditor); spawn prompts specify reactive behavior, locking roles into question-and-answer loops that preclude proactive code inspection; and the SendMessage primitive carries prose, encouraging abstract agreement (``\texttt{LBO=16}, confirm?'') over byte-level cross-checking.
Fixed-role multi-agent systems can therefore stall on debugging tasks that cross role boundaries.
\textsc{Evolving Agent Teams} addresses this class of failure: the supervisor diagnoses the stall as a \emph{knowledge gap} and introduces a Knowledge Integrator (Section~\ref{sec:mla-case-study}) that mandates cross-agent artifact reads.

\section{Details of the Additional Settings}
\label{app:generalization}

All four runs use the same supervisor prompt and harness as the main experiments, with no predefined roles. Each is a single run, so we report them as case studies rather than controlled comparisons; the matched-budget solo ablations are in Appendix~\ref{app:solo-ablations}. Costs are booked agent and supervisor API spend at published per-token prices.

\textbf{KDA prefill on AMD MI350X.}
Kimi Delta Attention~\citep{kimi2025linear} is a gated delta rule with a per-key-channel forget gate and a matrix-valued fp32 state; the kernel computes the chunked prefill from the raw projections (normalization, gates and recurrence). The task uses one MI350X (CDNA4) and the tensor-parallel shape of a production deployment ($H{=}12$ heads per GPU, $K{=}V{=}128$). The baseline is vLLM's hand-written fused HIP prefill kernels used in production~\citep{kwon2023vllm}; the starting kernel is a port of them (score $0.99$). The score is the arithmetic mean of baseline-over-candidate latency across 10 variable-length workloads (single sequences of 1K--64K tokens, batches from $4{\times}4$K to $64{\times}128$, and one mixed batch), with the cache flushed before each timed run. Correctness requires relative $L_2$ error $\le 2{\times}10^{-2}$ on outputs and final states; the team itself found that the evaluation's fast-decaying gates let state-handoff bugs pass, and added a slow-decay check that every subsequent kernel had to pass, along with a held-out shape set that is checked but not scored.
The run used Claude Opus~5.5 (medium effort) for all agents, lasted 73 iterations (186 hours), and booked \$3.0K. The final kernel scores $3.01\times$ (mean of five re-measurements; per-workload $2.09$--$4.63\times$) and is $7.3\times$ faster than the Triton kernel from Flash Linear Attention. Its main structural changes are a ``relay'' that fuses the prologue with the recurrence and passes the state between workgroups through the shared L2 cache, bf16/fp16 state handoffs and non-temporal loads, fp16 matrix operands ($+5.5\%$ and more accurate than bf16), and a ``twin route'' that feeds one 64-token recurrence step from two 32-token prologues.

\textbf{KDA prefill on B200.}
The task is the same operator on B200 (\texttt{sm\_100a}), scored as the geomean speedup over FlashKDA~\citep{flashkda2026}, a fused CUTLASS implementation, in a clean room with no web access. The team could read FlashKDA, the Triton KDA kernels of Flash Linear Attention, the CUTLASS/CuTe examples, FlashAttention-4~\citep{zadouri2026flashattention}, and a read-only snapshot of \textsc{FlashInfer}~\citep{ye2025flashinfer} at commit \texttt{a9e03bf2} (Aug.~13, 2026), which includes the SM100 KDA prefill kernels that CAKE~\citep{cake2026} contributed to \textsc{FlashInfer}; the team ported these in the first iteration ($1.96\times$).
The score covers six workloads (8{,}192 tokens, $H \in \{64, 96\}$, fixed and variable-length layouts).
Iterations 1--5 used Claude Opus~5.5 and later iterations Claude Sonnet~5.5. \emph{Comparison with KDA$^2$.} KDA$^2$~\citep{kda2_2026} reports $2.96\times$ over FlashKDA on B300 on the same six workloads. We recompiled the team's final kernel for B300 (\texttt{sm\_103a}, no code change) and scored it and KDA$^2$'s released kernels with the same harness on the same B300: $3.54\times$ for ours, against $3.02\times$ (TIRx), $2.98\times$ (PTX), and $2.85\times$ (CuTe-DSL) for KDA$^2$. On B200, the same harness gives $3.44\times$, against $2.91\times$ (TIRx) and $2.74\times$ (CuTe-DSL); the PTX kernels run only on B300. Because the run starts from a port of an existing kernel, it measures improvement over a strong start rather than from scratch.

\textbf{MoE expert-offload serving on Apple silicon.}
The task is to make a Metal port of the FreeToken MoE serving system~\citep{yang2026freetoken} serve Qwen3.8-Flash-Next at 4 bits (about 79\,GB of weights plus a 32\,GB per-layer embedding table; 512 routed experts per layer, top-10 plus a shared expert) as fast as possible on a 48\,GB M4 Pro Mac, streaming experts that do not fit in memory from SSD through an offload cache. The model and math must stay the same: no expert skipping or pruning, no lower top-$k$, no fewer bits. Gates check top-1 agreement with the naive engine's greedy predictions, KL divergence, peak memory ($\le 37$\,GiB), swap growth, and load time, and a candidate that fails any gate scores 0. The score is the mean over workload terms of baseline time over candidate time; the workload replays sessions recorded from a real coding-agent loop, teacher-forced from a cold page cache, plus a batch-1 decode benchmark. All agents used Claude Opus~5.5, with web search and fetch tools disabled.
The supervisor resized the team from one agent to three and finally to two, and re-cut its roles six times: by file type (I/O vs.\ compute), by bottleneck (speculation, cache, offload), by time slice (decode vs.\ prefill), and by measured gap (GPU time per token). Main changes include a victim cache for evicted experts, compact GPU expert slots with shared palette indices, (scale, bias) palette compression that cuts expert reads by 7\%, and admission rather than eviction as the residency lever, after a cache role found that most decode misses had already been read during the preceding prefill. The final engine decodes at $27.3$ tokens/s on the batch-1 decode benchmark, against $13.4$ for the naive engine. In Figure~\ref{fig:new-domains}c, iterations 3--6 have no valid evaluation, and each iteration's cost is its team cost scaled so that the total matches the booked \$1.0K, which includes the supervisor.

\textbf{RoboDojo.}
RoboDojo~\citep{chen2026robodojo} has 42 dual-arm manipulation tasks in Isaac Sim, grouped into five dimensions (generalization, memory, precision, long-horizon, open-ended); the leaderboard score is the mean of the five dimension means of per-task partial-credit scores (0--100). The team edits a Python controller package with per-task skill modules. At run time the policy may call the frozen $\pi_{0.5}$ checkpoint~\citep{black2025pi05} and a runtime LLM (Claude, at most 20 calls per episode); inputs are three RGB cameras, joint states, and the instruction, and no model training is allowed. The agents' outbound network is limited to the Anthropic API and the evaluation fleet, and other leaderboard entries' policy code is removed from the hosts. Agents develop on seed~1; seeds~0 and~2 are held out and evaluated only by the operator, and agents never see those scores. Iterations 1--36 optimize a 10-task subset; iterations 37--51 use all 42 tasks.
The supervisor and all agents used Claude Sonnet~5.5, and the run booked \$472 through iteration~50 (excluding runtime-LLM calls). The iteration-50 policy scores $62.1$ held-out (50 episodes per task on each held-out seed) and $62.6$ on dev.

\section{Matched-Budget Solo Ablations}
\label{app:solo-ablations}

On two of the additional settings we compare the team against a single Claude Code agent with the same model at matched spend.
On MI350X, from the same starting commit and task statement, at \$300 the solo agent (Opus~5.5, medium effort) reaches $1.65\times$, while the team reaches $2.13\times$.
On the offload task, starting from the team's $3.2\times$ engine, the two are close at \$200 ($1.047\times$ for solo vs.\ $1.056\times$, averaged over the dev and two held-out workloads), but within the same 20-hour window the solo agent progresses more slowly and spends only \$200, while the team reaches $1.083\times$ at \$509.

\begin{figure}[h]
\centering
\includegraphics[width=0.48\textwidth]{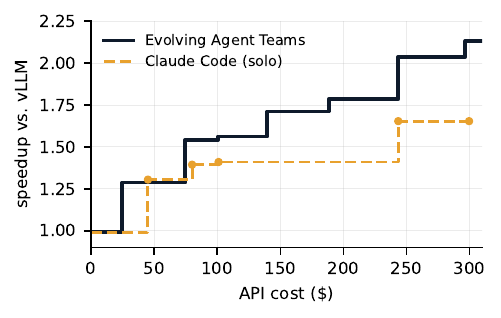}
\caption{KDA prefill on MI350X over the first \$300.}
\label{fig:amd-solo}
\end{figure}

\textbf{KDA prefill on AMD MI350X.}
The solo run starts from the team's starting commit and task statement. Figure~\ref{fig:amd-solo} compares the team and the solo run at matched spend.

\textbf{MoE expert-offload serving, from an optimized start.}
A second run starts both the team and a solo agent from the final engine of the run in Appendix~\ref{app:generalization} (about $3.2\times$ naive), each on two identical Macs, both shown a \$1{,}000 budget, and both stopped at the same 20-hour deadline. Table~\ref{tab:ftm-solo} reports all versions re-measured afterwards in the same batch on the same machines, on the dev workload and on two held-out workloads: (A) a long prefix-reuse session and (B) four real Claude Code sessions. At \$200 the team and the solo agent are close; the solo agent leads on the workload it optimized and the team on both held-out workloads. The solo agent's later changes are mostly small parameter tweaks (e.g., the half-life of the expert-cache eviction policy), which help the dev workload more than the held-out ones. The solo agent progressed more slowly and spent \$200 in the 20 hours, while the team kept improving through \$509 (decode lookahead hints, then exact multi-row verification enabling speculative decoding, then a larger victim cache), and its final engine is the fastest on every workload. Measured together on the same Macs, it decodes at $30.3$ tokens/s on the batch-1 decode benchmark, against $26.7$ for the starting commit and $28.2$ for the solo agent's final engine.

\begin{table}[h]
\centering
\caption{MoE offload serving, second run: speedup over the shared starting engine, re-measured in the same batch on the same machines (macOS 26.7). ``All'' is the equal-weight mean of the three workloads. Rows are labeled by wall-clock time since the start; both runs stopped at the same 20-hour deadline, and the team reached the solo run's final cost (\$200) after about 8 hours.}
\label{tab:ftm-solo}
\vspace{3pt}
\small
\setlength{\tabcolsep}{4.8pt}
\begin{tabular}{lccccc}
\toprule
\textbf{Version} & \textbf{Cost} & \textbf{Dev} & \textbf{Held-out A} & \textbf{Held-out B} & \textbf{All} \\
\midrule
Claude Code (solo), 20\,h & \$200 & $1.060$ & $1.036$ & $1.045$ & $1.047$ \\
\textsc{Evolving Agent Teams}, 8\,h & \$200 & $1.050$ & $1.062$ & $1.055$ & $1.056$ \\
\textsc{Evolving Agent Teams}, 20\,h & \$509 & $\mathbf{1.098}$ & $\mathbf{1.080}$ & $\mathbf{1.072}$ & $\mathbf{1.083}$ \\
\bottomrule
\end{tabular}
\end{table}

\end{document}